\documentclass[10pt,twocolumn,letterpaper]{article}
\pdfoutput = 1
\usepackage{iccv}
\usepackage{times}
\usepackage{epsfig}
\usepackage{graphicx}
\usepackage{amsmath}
\usepackage{amssymb}
\usepackage{textcomp}
\usepackage[pagebackref=true,breaklinks=true,letterpaper=true,colorlinks,bookmarks=false]{hyperref}

\iccvfinalcopy % *** Uncomment this line for the final submission

\def\iccvPaperID{80} % *** Enter the ICCV Paper ID here

\ificcvfinal\fi
\begin{document}

%%%%%%
% Title %
%%%%%%
\title{TurkerGaze: Crowdsourcing Saliency with Webcam based Eye Tracking}

\author{\normalsize 
Pingmei Xu~~~~Krista A Ehinger$^\star$~~~~Yinda Zhang~~~~Adam Finkelstein~~~~Sanjeev R. Kulkarni~~~~Jianxiong Xiao
\\\normalsize Princeton University~~~~~~~$^\star$Harvard Medical School and Brigham \& Women's Hospital~~~~~~~}
\maketitle

%%%%%%%%%
% Abstract %
%%%%%%%%%

\begin{abstract}
Traditional eye tracking requires specialized hardware, % in a lab, % with careful calibration,
which means collecting gaze data from many observers is expensive, tedious and slow.
Therefore, existing saliency prediction datasets are order-of-magnitudes 
smaller than typical datasets for other vision recognition tasks.
The small size of these datasets
limits the potential for training data intensive algorithms, and 
causes overfitting in benchmark evaluation.
To address this deficiency, this paper introduces
a webcam-based gaze tracking system that supports
large-scale, crowdsourced eye tracking
deployed on Amazon Mechanical Turk (AMTurk).
By a combination of careful algorithm and gaming protocol design,
our system obtains eye tracking data for saliency 
prediction comparable to data gathered in a traditional lab setting,
with relatively lower cost and less effort on the part of the researchers.
Using this tool, we build a saliency dataset 
for a large number of natural images.
We will open-source our tool and provide a web server
where researchers can upload their images to get eye tracking results 
from AMTurk.
\end{abstract}

%%%%%%%%%%%%
% Introduction %
%%%%%%%%%%%%
\section{Introduction}
\label{sec:intro}

An understanding of human visual attention is essential to many applications in computer vision, computer graphics, computational photography, psychology, sociology, and human-computer interaction \cite{Santella06,Ji04}. Eye movements provide a rich source of information into real-time human visual attention and cognition \cite{Borji13_2}, and the development of gaze prediction models is of significant interest in computer vision for many years. However, gaze prediction is difficult because the complex interplay between the visual stimulus, the task, and prior knowledge of the visual world which determines eye movements is not yet fully understood \cite{Butko10}. Recently, the availability of eye tracking devices and machine learning techniques opens up the possibility of learning effective models directly from eye tracking datasets.  \cite{Papadopoulos14,Mathe13,Kummerer14} demonstrated that eye tracking datasets on a larger scale are critical to leverage the power of machine learning techniques and greatly improve the performance of visual attention models. However, building large databases of eye tracking data has traditionally been difficult, typically requiring bringing people into a lab where they can be tracked individually using commercial eyetracking equipment. This process is time-consuming, expensive, and not easily scaled up for larger groups of subjects.

\begin{figure}[t]
\begin{center}
%\fbox{\rule{0pt}{2in} \rule{0.9\linewidth}{0pt}}
\includegraphics[width=\linewidth]{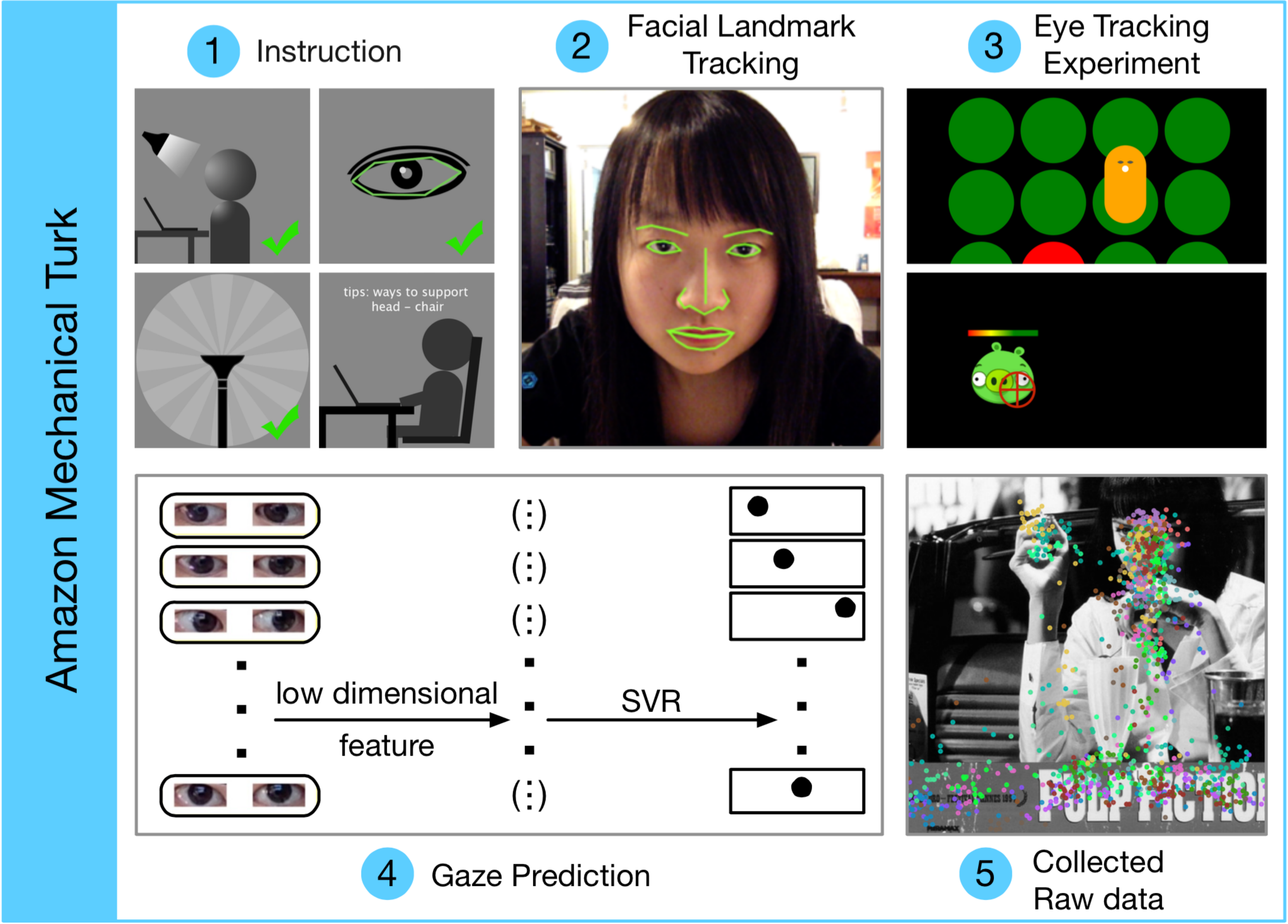}
\end{center}

\vspace{-3.5mm}
\caption{\textbf{We propose a webcam-based eye tracking system to collect saliency data on a large-scale.} By packaging the eye tracking experiment into a carefully designed web game, we are able to collect good quality gaze data on crowdsourcing platforms such as AMTurk.}
\label{fig:system_overview}

\vspace{-4mm}
\end{figure}

In this paper, we address this deficiency by designing a webcam-based eye tracking game for large-scale crowdsourcing on Amazon Mechanical Turk (AMTurk) as shown in Figure~\ref{fig:system_overview}. 
We designed our system around the following criteria: ubiquitous hardware, ease of set-up, sufficient quality for saliency data collection, and real-time performance. 
The challenges in developing such a system include arbitrary lighting and other environmental conditions for the subject, and the need for performing the bulk of the tracking and related computation on limited and varying hardware available on the subject's computer.

The contributions of this work include building a robust tracking system that can run in a browser on the remote computer, design and evaluation of two game scenarios that motivate the subjects to provide good gaze data, a framework that supports experiments of this kinds, and a large database of gaze data. We also show that our system obtains eye tracking data from AMTurk with satisfactory accuracy compared with the data gathered in a traditional lab setting, but at a lower cost and with little effort from the researchers.

\section{Related work}
In this section, we briefly review existing computer vision algorithms for eye tracking and some recent methods for collecting crowdsourced saliency data. We also describe the exisiting eye tracking datasets and stimuli.

\subsection{Computer vision based eye tracking}
There exist two main categories of computer vision based gaze prediction methods: feature-based and appearance-based methods \cite{Hansen10}. Feature-based methods extract small scale eye features such as iris contour \cite{Wang03}, corneal infrared reflections and pupil center \cite{Valenti08} from high resolution infrared imaging, and use eyeball modeling \cite{Guestrin06} or geometric method \cite{Yoo05} to estimate the gaze direction. This approach requires a well-controlled environment and special devices such as infrared/zoom-in cameras and pan-tilt unit \cite{Morimoto05}. In addition, the accuracy depends heavily on the system calibration. On the other hand, appearance-based methods use the entire eye image as a high-dimensional input to a machine learning framework to train a model for gaze prediction, so the image from a standard camera or webcam is sufficient.

Several appearance-based methods that assume a fixed head pose have been proposed. Tan \etal \cite{Tan02} used a manifold model with densely sampled spline (252 calibration points). Williams \etal \cite{William06} proposed a Gaussian process regression to reduce the number of required samples. Lu \etal \cite{Lu11} introduced an adaptive linear regression method which only requires sparse samples (9-33 calibration points) and achieves a fairly high accuracy. However, as mentioned in Section \ref{sec:intro}, small head movements inevitably lead to a drift of gaze prediction during the experiment. A few solutions have been reported, but most require additional calibration. Lu \etal \cite{Lu11_2} suggested learning a compensation for the head pose change by recording a video clip during calibration, but this requires nearly 100 training images from different head poses. Lu \etal \cite{Lu12} proposed using synthetic training images for varying head poses, but this still requires users to move their heads vertically/horizontally to capture more reference images after gazing at 33 calibration points. These tedious additional calibration procedure makes a direct application of these methods to crowdsourcing very difficult.

\subsection{Crowdsourced saliency}
The possibility of crowdsourcing human attention data has been an increasing popular topic in recent years, and a few methods exist. The current methods are alternatives to standard eyetracking: they do not collect real-time gaze data, but instead collect proxy data which can be used to train saliency models.

Rudoy \etal \cite{Rudoy12} asked participants to watch a clip of video, immediately followed by a briefly-displayed character chart, and asked participants to self-report their gaze location by giving the letter and number symbols which they had seen most clearly. Although the accuracy of this approach is comparable to data collected in-lab by an eye tracker, only a single fixation can be obtained in each self-report. Frame-by-frame eye tracking of an entire video clip seems infeasible, and it is unclear if this system could be used to collect scan paths on static images.

Recently, Jiang \etal \cite{Jiang15} designed a mouse-contingent paradigm to record viewing behavior. They demonstrated that the mouse-tracking saliency maps are similar to eye tracking saliency maps based on shuffled AUC (sAUC) \cite{Zhang08}. However, mouse movements are much slower than eye movements, and the gaze pattern of individual viewers in mouse and eye tracking systems are qualitatively different. It is not clear that if the "fixations" generated from mouse tracking match those obtained from standard eye tracking (the authors propose extracting fixations by excluding half samples with high mouse-moving velocity, but this leaves up to 100 fixations per second, far different from the approximately 3 fixations/second observed in standard eye tracking data). Furthermore, since mouse movements are slower than eye movements, it's unclear if this approach will work for video eye tracking, or if it can be used in psychophysics experiments which require rapid responses.

Finally, general image labelling tools such as LabelMe \cite{Russell08} can be used to identify salient objects and regions in images by crowdsourcing. This provides valuable data for developing saliency models, but is not a substitute for real-time eye tracking.

\subsection{Eye tracking datasets}
In-lab eye tracking has been used to create data sets of fixations on images and videos. The data sets differ in several key parameters \cite{Borji13}, including: the number and style of images/video chosen, the number of subjects, the number of views per image, the subject’s distance from the screen, eye trackers, the exact task the subjects were given (free viewing \cite{Kootstra11}, object search \cite{Papadopoulos14}, person search \cite{Ehinger09}, image rating \cite{Wang10} or memory task \cite{Judd09}), but each helps us understand where people actually look and can be used to measure performance of saliency models. The majority of eye tracking data is on static images. The most common task is free viewing which consists of participants simply viewing images (for 2-15 seconds) or short videos clips \cite{Mathe14} without no particular task in mind. 

Two widely-used image datasets are the Judd dataset which contains 1003 natural images free-viewed by 15 subjects each \cite{Judd09} and the NUSEF dataset \cite{Ramanathan12} which includes 758 (emotion evoking) images free-viewed by 25 subjects each. There are other datasets focused on specific domains: PASCAL-S \cite{Li14} offers both fixation and salient object segmentation; OSIE \cite{Xu14} features in multiple dominant objects in each image; and the MIT Low Resolution dataset \cite{Judd11} contains low resolution images. Building these databases typically involves a trade-off between number of images and number of human observers per image. The POET database \cite{Papadopoulos14} is currently the largest database in terms of images (6270), but each image is only viewed by 5 subjects performing a visual search task for one second.

%%%%%%%%%%%%
% Our system  %
%%%%%%%%%%%%

\section{Large-scale crowd-sourcing eye tracking}
In this section we first review main challenges for collecting a large eye tracking dataset via crowdsourcing. Next, we describe our platform and the experiment setup.
\subsection{Design considerations}
\label{sec:challenges}
\textbf{Experiment environment.} Unlike a lab setting, the crowdsourced worker's environment is uncontrolled. Relevant factors that need to be considered include lighting, head pose, the subject’s distance from the screen, etc.

\textbf{Hardware.} Although our system only requires a HD webcam and a modern web browser, there are still a number of workers who cannot meet the hardware requirements due to a low computational speed of the web browser or a low sample rate of a webcam. These workers must be detected and screened out.

\textbf{Software and real-time performance.} Although there is some previous work on webcam based gaze prediction, none of these approaches have been implemented in a web browser. We desire a web-ased system that can make eye predictions in real time.  This would allow us to: 1) provide online feedback, which is critical to attract and engage subjects; 2) avoid making a video recording of our participants, which protects their privacy. Since modern web browsers have limited computational resources, this requires a simple algorithm and 
an efficient implementation.

\textbf{Head movement.} In a traditional lab set-up, a chin rest is used to fix the position of a participant's head, but this is inpractical to enforce during crowdsourcing. Gaze prediction algorithms that work well with a fixed head perform significantly worse when the head is not stationary. This is because head movements cause the appearance of the eye regions to change so drastically that they cannot be matched to the data recorded during calibration even if they correspond to the same gaze position \cite{Sugano08}.

\textbf{Lack of attention.} Whether the eye tracking participants can focus with their full attention on the task directly determines the quality of the data gathered. This is a real challenge for crowdsourcing. Therefore, users must be given clear instructions and the experiment procedure must be designed to be fun and engaging.

\subsection{Webcam based gaze prediction algorithm}
\label{sec:gaze_prediction}

Based on these considerations, this section introduces our method for eye tracking, while Section \ref{sec:game_design} proposes a gaze data collection procedure in the form of a game.

\textbf{Facial landmark tracking \& eye region extraction. } 
To track facial landmarks we used the JavaScript package \emph{clmtrackr} \cite{clmtracker}, which implements constrained local models fit via regularized landmark mean-shift, as described in \cite{Saragih11}.
The estimation of facial landmarks suffers from visible jittering due to small head movements and optimization challenges, so we further stabilized these landmark locations by temporal smoothing using a Kalman filter.
From each video frame, we used the location of eye corners and eyelids as reference points for rectangular eye region alignment and extraction. We set a minimum value of the size of the eye image and of the average intensity to ensure that: (1) the subject is not too close or too far from the camera; (2) the resolution of eye image is sufficient; and (3) the subject's eyes are well lit.

\textbf{Calibration.}
Our system trained a user-task specific model for gaze prediction, which means that to obtain training data, in each experiment session the user was requested to gaze at certain positions (calibration points, shown in Figure \ref{fig:game_setting}) on the screen. Users stared at each point for 1 second; to ensure a stable eye position we used only the last several frames from each point to train the model. For the eye images corresponding to each calibration point, we computed the Zero mean Normalized Cross-Correlation (ZNCC) between each image and their average. Eye images with low ZNCC value were classified as blinks and discarded. We used the remaining frames to train the gaze prediction model. In our experiments, we set the threshold to be 95\% of the largest ZNCC.

\textbf{Gaze prediction.}
A typical size of the eye images is $40 \times 70$ pixels, so the dimensionality of a pixel-wise feature vector can reach into thousands of dimensions. With sparse samples of gaze positions for training, a high-dimensional feature can easily overfit, so instead we adopted a similar approach as Lu~\etal~\cite{Lu11}. We rescaled each eye image to $6\times10$ and then performed histogram normalization, resulting in a 120-D appearance feature vector for both eyes.

To achieve real-time performance for online applications, we use ridge regression (RR) to map the feature vector to a 2-D location on the screen. In a subsequent offline step after the experiment is finished, we retrain the model using support vector regression (SVR). For our experiments we use a Gaussian kernel and C-SVC \cite{Chang11}.

\subsection{Game design for crowdsourcing}
\label{sec:game_design}
As discussed in Section \ref{sec:challenges} the uncontrolled setting through the AMTurk platform makes the data collection of eye tracking with webcams an extremely difficult task due to issues such as bad lighting, large head movement, lack of attention to both calibration and assigned tasks, and so forth. 
%In this section, we will discuss in detail how we tackled these various problems. 
Due to these deployment issues, the design and implementation of both the user experience and the user interface are not trivial but are crucial to this task. 

\begin{figure}[t]
	\begin{center}
		%\fbox{\rule{0pt}{2in} \rule{0.9\linewidth}{0pt}}
		\includegraphics[width=\linewidth]{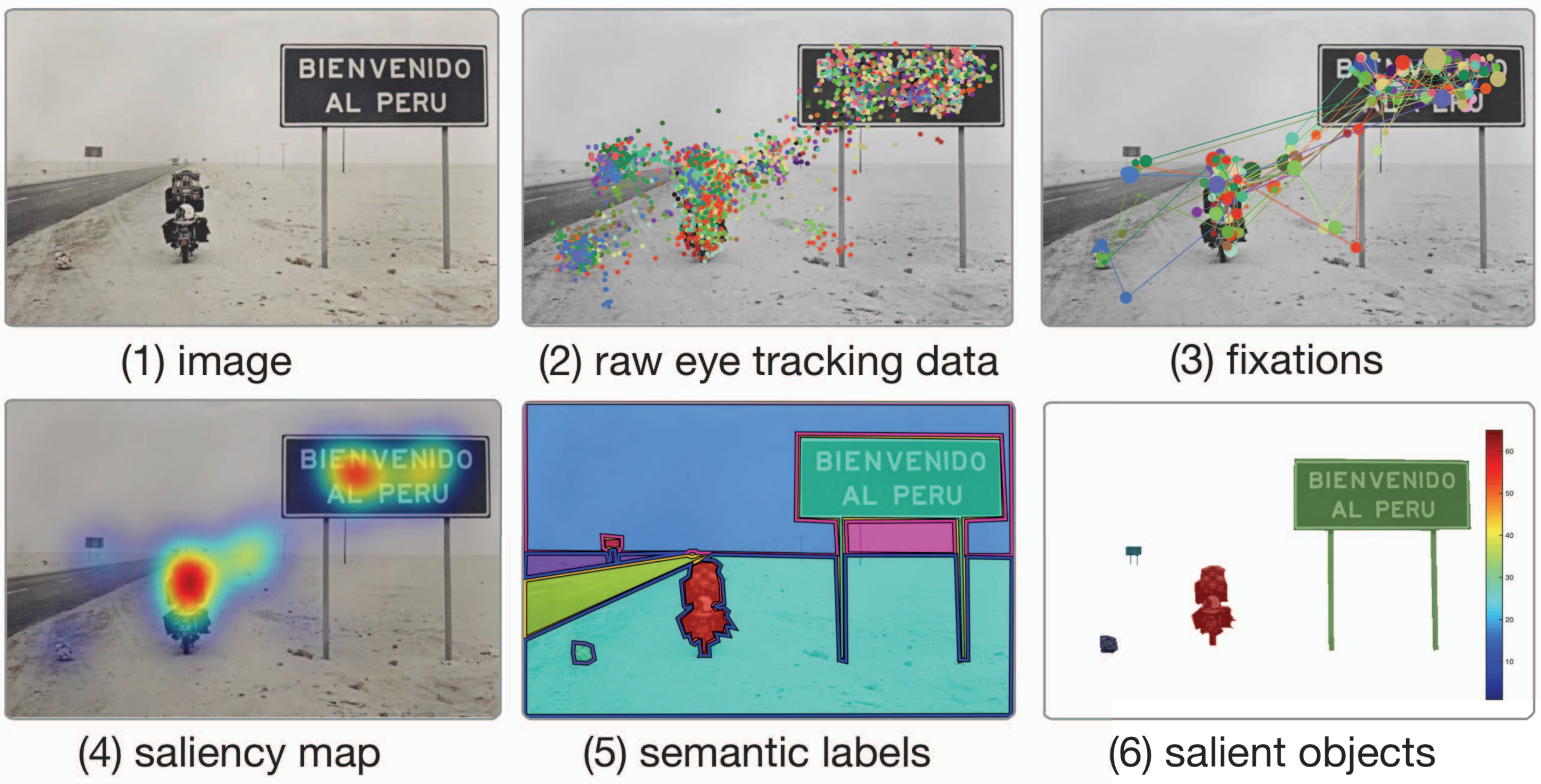}
	\end{center}
	
	\vspace{-4mm}	
	\caption{\textbf{An example of the saliency data obtained by our system.} In a free-viewing task, users were shown selected images from the SUN database with fully annotated objects. From the raw eye tracking data, we used the proposed algorithm to estimate fixations and then computed the saliency map. This map could be used to evaluate the saliency of individual objects in the image.}
	\vspace{-2mm}  
	\label{fig:free_view}
\end{figure}

\subsubsection{Quality control for prediction accuracy}
Our early prototypes showed that by using the procedure mentioned in Section \ref{sec:gaze_prediction} with only calibration and image viewing, the majority of submitted work did not give a satisfactory result. After scrutinizing each video of eye images, we discovered that the gaze prediction may fail due to a variety of reasons such as head movement, head pose, bad lighting, unsupported browser features, low computational speed, and low memory capacity. The primary cause of failure is head movement, which is our main concern when making design decisions. Instead of explicitly allowing free head pose as some previous works \cite{Lu11_2}, we instruct the user to stay as still as possible by supporting their head on their arms, books or the back of a chair. In addition, we proposed two design possibilities to deal with this issue: 1) check and recalibrate frequently; 2) use an extremely short task.

\begin{figure*}[t]
	\begin{center}
		\includegraphics[width=1\linewidth]{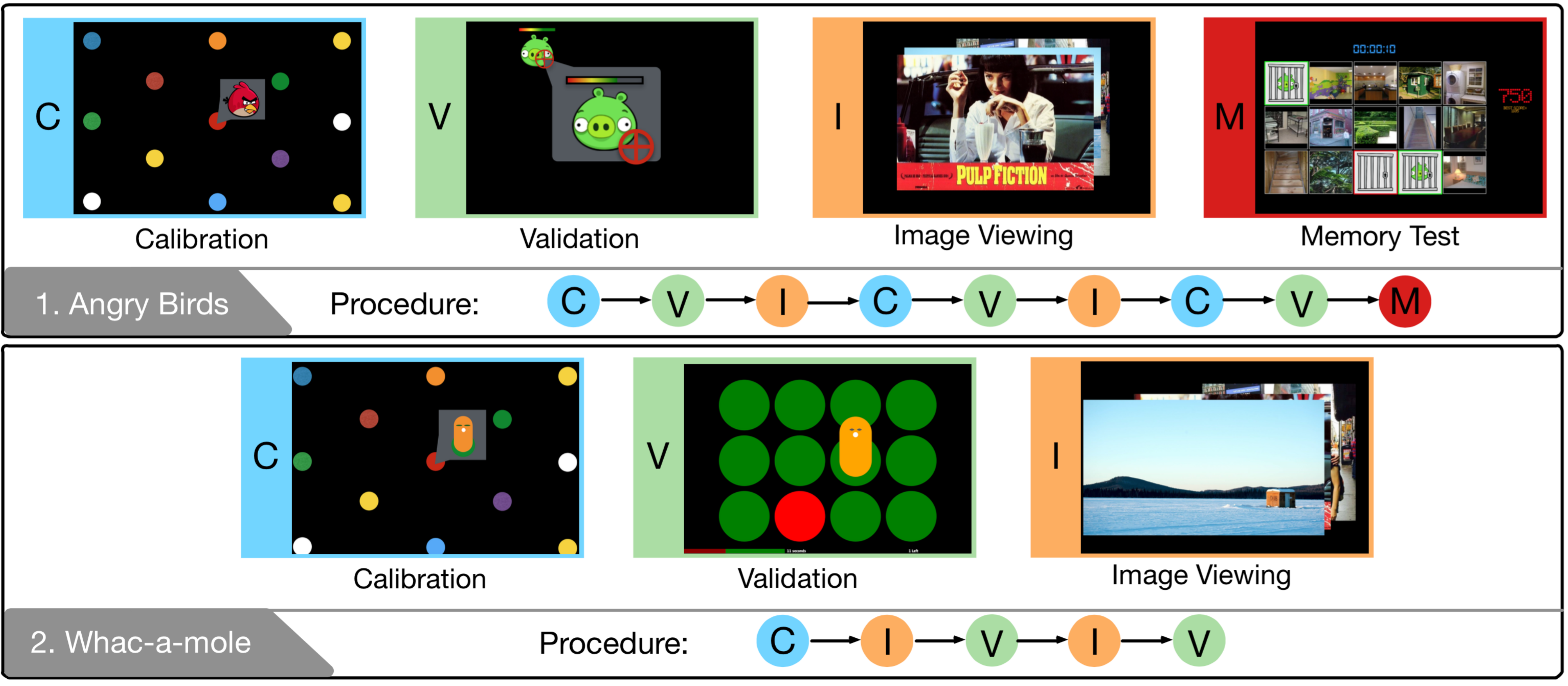}
	\end{center}
	
	\vspace{-3mm}
	\caption{\textbf{The procedure for an eye tracking experiment.} 13-point calibration and validation were performed at the start of each session. During the calibration, we trained a user-task specific model for gaze prediction. For validation, we displayed the online gaze prediction results in the context of a game to make sure that the tracking was functioning properly. Our system supports various stimuli (such as image or video) and various tasks (such as free viewing, image memory, or object search).}
	\vspace{-2mm}   
	\label{fig:game_setting}
\end{figure*}

As shown in Figure \ref{fig:game_setting}, in each experiment, we provided multiple (N) sessions of calibration sandwiching N-1 sessions of photo viewing. At the end of each session, a new model was trained from all the data collected to that point, to give an updated online gaze prediction. After each calibration session, we inserted a validation stage when the user was asked to gaze at certain positions, and we checked the online prediction to ensure high-quality tracking results. 

Occasionally subjects missed a few calibration points, which results in an inaccurate model that substantially deviates from the average case. In order to reject outliers, we partitioned the training samples into N folds based on their calibration session and conducted N-fold cross validation. Then we grouped samples corresponding to the same gaze position across all folds and discarded those with large leave-one-out error. The dropout rate was set to $100/N\%$, which allowed to reject at most the entire set of samples corresponding to a certain calibration point. Finally, we used all selected frames to fit a refined model by SVR.

An alternative to frequent recalibration is to keep tasks short (under a minute); in this case, it is less likely that the user will unintentionally move their head. For example, in a free viewing task, a typical viewing time per image is 3-5 seconds, so it is easy to make the task short by only showing a small number of images. In this case, we used only one calibration but still add a validation stage to detect failure cases.

Next, we briefly discuss how we dealt with other possible causes of failures. 1) For head pose, we instructed the user face forward to the screen. Before starting each experiment, we estimated the approximate jaw, pitch and yaw based on the position of facial landmarks. If we detected that the face pose was not frontal, we asked the user to change the pose accordingly. During the experiment, the system would restart if a large head movement was detected. 2) For lighting, we suggested the user sit in a well-lit area with their face lit from the front. We computed the average intensity of eye regions and asked the user to adjust lighting conditions if the average intensity was below a threshold. 3) For hardware support, we conducted a system check to test the resolution of the webcam (at least $1080 \times 720 p$) and the frame rate of recording (at least $25$ fps) when the facial landmark tracking was running. Only individuals who passed the system check could continue the experiment.

\subsubsection{Quality control for attention}
For free viewing tasks, we provided a memory test to motivate the subjects to pay attention to the images as in \cite{Judd09}: we showed 15 images arranged on a grid and asked them to indicate which $M$ images they had just seen. We chosed $M$ to be 1 -- 3 depending on the experiment configuration. 

We observed some cases where subjects did not appear to be viewing images normally, even though this was emphasized in the instruction. Therefore, without screening the results, the average AMTurk saliency map has a stronger center bias than the in-lab data. To deal with this issue, we added images from Judd dataset as a benchmark to evaluate subjects’ performance. These images were selected to have low entropy saliency maps with one salient object far from the center.

Eye tracking is not an everyday experience for most of the workers on AMTurk, so proper feedback to assure the subjects that the eye tracking is functioning well is important to get them engaged into the experiment. After each calibration session, we displayed the online prediction in the context of a game, which we will discuss more details in Section \ref{sec:game_interface}.

\subsubsection{Game interface}
\label{sec:game_interface}
We designed two game interfaces for collecting eye tracking data on a crowdsourcing platform as shown in Figure \ref{fig:game_setting}. The basic idea is to fit our experiment into the storyline of well known games and provide real-time feedback to 1)  leverage the impact of these games to attract more workers; 2) encourage them to work more carefully; and 3) integrate the quality control to the experiment naturally.

\textbf{Theme one:} Angry Birds \cite{Angrybird}. In the original video game, players use a slingshot to launch birds at pigs stationed in or around various structures with the goal of destroying all the pigs on the playing field. For this experiment, we used a calibration pattern with 13 points. During the calibration, instead of showing red dots on the screen as in a traditional eye tracker, we popped bullets (birds in Figure \ref{fig:game_setting}) sequentially according to the calibration pattern with sound effect and asked the subject to carefully check each one of them in order to train a powerful gun controlled by gaze. Next, during the validation, we showed a target (pig in Figure \ref{fig:game_setting}) at a random location on the screen, displayed the gaze prediction results obtained from the online model as a crosshair, and asked the subject to use their gaze to shoot the target. If they successfully focused their gaze on the target for a sufficient number of time, they were able to ``kill'' the target, earn bonus points, and move on to the next stage. We used this theme for our longer experiments, so a multistage calibration and memory test were also used.

\textbf{Theme two:} Whac-A-Mole \cite{Whac-a-mole}. In the original mechanical game, moles will begin to pop up from their holes at random, and players use a mallet to force them back by hitting them on the head. For this interface, we used a simpler calibration pattern with 9 points. After collecting mallets during calibration, we asked the user to stare at the moles to force them back into their holes. We used this theme for short experiments, with less number of calibration sessions and less number images for free viewing. 

We included these two interfaces to maximize the usage of the crowdsourcing resource. The first theme allows us to provide more accurate online gaze prediction, but it requires subjects to keep their head very still and carefully maintain their gaze on the calibration and validation points. Subjects who failed to control the crosshairs to kill this pigs in time would get frustrated and leave the experiment, so the higher accuracy came at a cost of slower data collection. Therefore, we also used the Whac-a-Mole task which is extremely short (less than 30 seconds) and easy to play. The validation step in this task is easier for subjects to pass -- instead of displaying the exact online gaze prediction, we only use the rough location to mark which mole hole they are targeting. This gives subjects feedback that the tracking is working and lets them feel that they have control of game. After they finish the experiment, we then refine the prediction results offline. Although in each experiment we collected less number of images, it drastically increased the rate of successful completion of our task as well as the popularity of our tasks among workers, and we were still able to use the benchmark images to screen results. In practice, we published both games on AMTurk and let the workers choose which one they would like to do.

\begin{figure}[t]
	\begin{center}
		%\fbox{\rule{0pt}{2in} \rule{0.9\linewidth}{0pt}}
		\includegraphics[width=\linewidth]{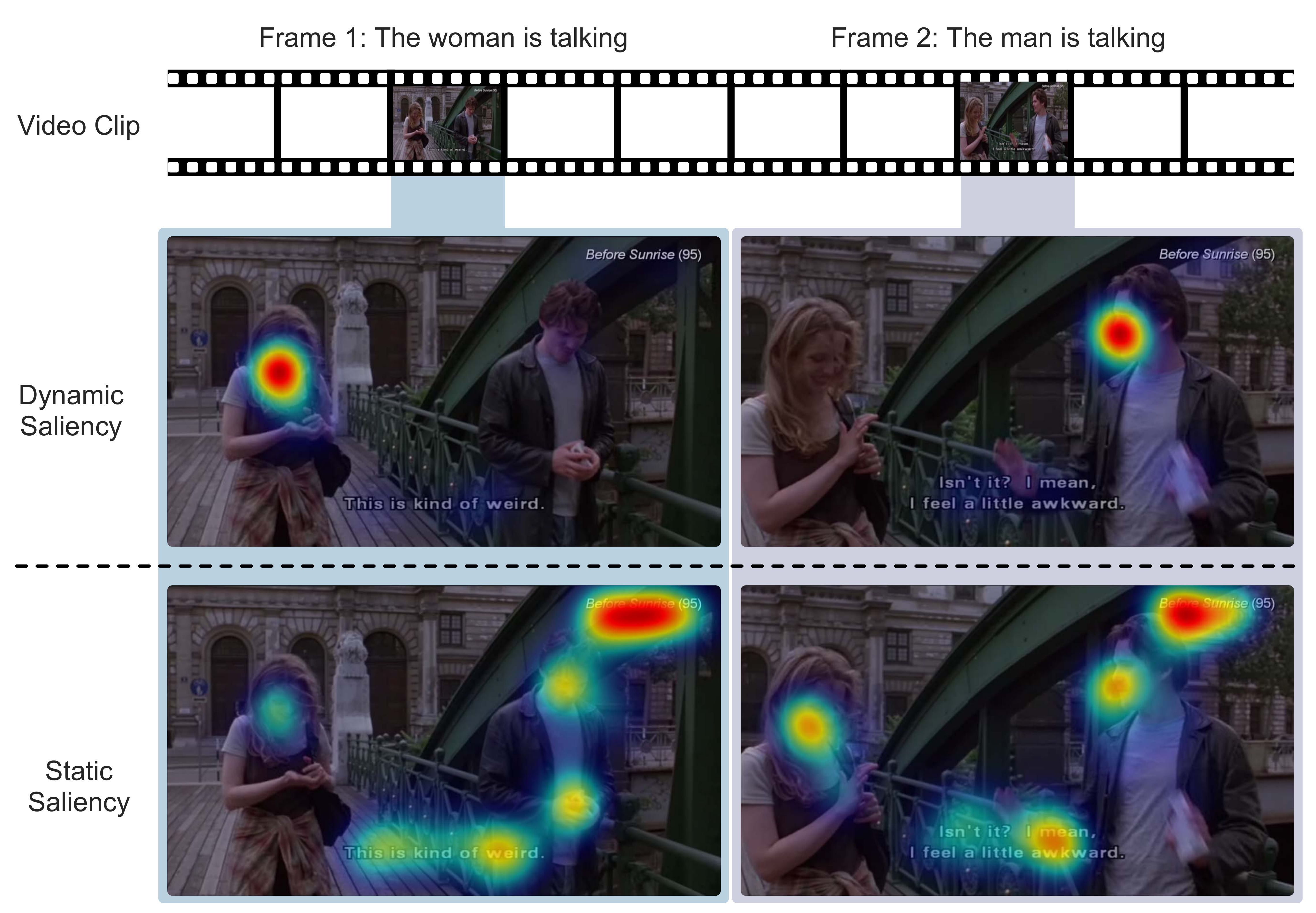}
	\end{center}
	
	\vspace{-4.5mm}
	\caption{\textbf{An example of the dynamic saliency maps for video clips collected by our system.} Saliency in video is highly dependent on the context of the previous frames, so dynamic saliency maps look very different from the static saliency maps obtained for the same frames shown in isolation as part of a free-viewing task.}
	\vspace{-1mm}   
	\label{fig:video_saliency}
\end{figure}

\subsection{Experiment setup}
\textbf{Stimuli.} The bulk of our experiments focus on collecting gaze data for natural images (Figure \ref{fig:free_view}). However, our system easily generalizes to other stimuli such as video (Figure~\ref{fig:video_saliency}) and panoramic imagery (Figure~\ref{fig:PanoSal}). More examples are shown in supplementary material.

\textbf{Protocol.} We implemented our system in Javascript and it runs by a  web browser in full screen mode. In each photo viewing session, we showed 4-10 images depending on the worker’s performance in previous experiments in the ‘Angry Bird’ interface and 1-2 images per session in the ‘Whac-a-mole’ interface. We scaled images to be as large as possible in the full screen mode while keeping their aspect ratio. Each image was presented for 3.5 seconds separated by 0.5 seconds of a black background with a blue fixation cross in the center of the screen and a number indicating the number of images left.

\textbf{Data processing.} A standard method to extract fixations is to detect saccade onset using a velocity and/or acceleration threshold; other methods using temporal or spatial thresholds have been propsed (\cite{Salvucci2000}). However, acceleration/velocity thresholds do not reliably detect saccades in our system because of the relatively low sample rate, so instead we used meanshift clustering in the spatio-temporal domain (i.e., [$x$, $y$, $t$] with a kernel of size [32, 32, 200]). We treat clusters with at least 2 samples as fixations and use cluster center as the fixation $x,y$ location. For analysis, we discarded the first fixation from each scan path to avoid adding trivial information from the initial center fixation as~\cite{Judd09}. 

\begin{figure}[t]
	\centering
	\includegraphics[width=\linewidth]{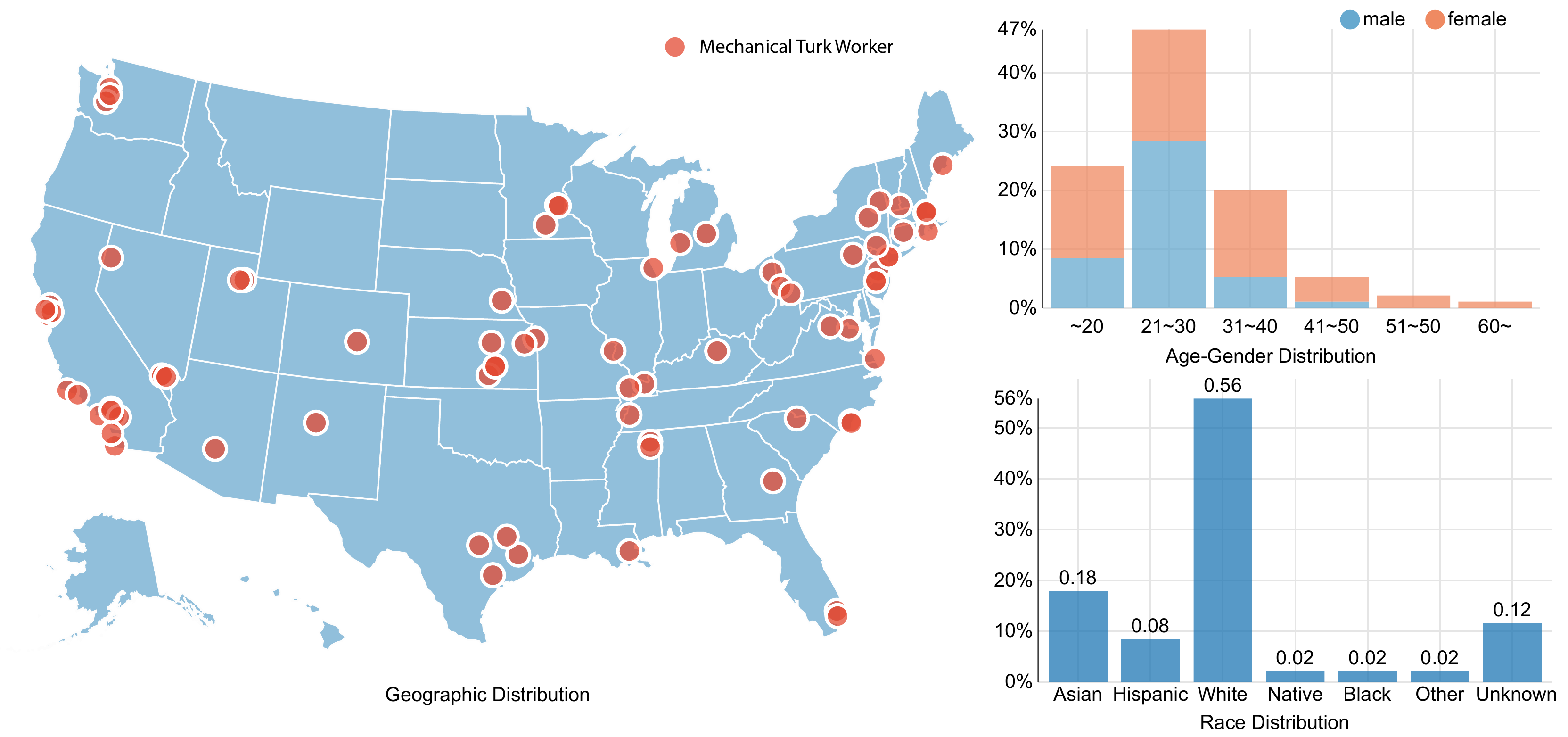}
	
	\vspace{-1mm}
	\caption{{\bf Statistics of basic information for workers from AMTurk.} The data is obtained from 200 randomly-selected workers  who participated our eye tracking experiment.}
	\label{fig:TurkerStatistics}
	\vspace{-2mm}
\end{figure}

\begin{figure}[t]
	\includegraphics[width=0.495\linewidth]{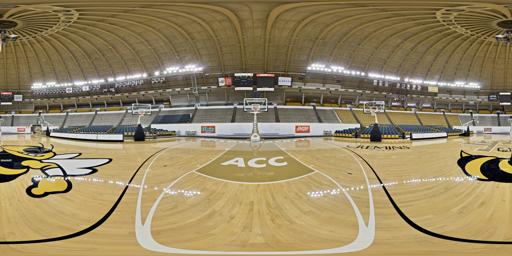}~%
	\includegraphics[width=0.495\linewidth]{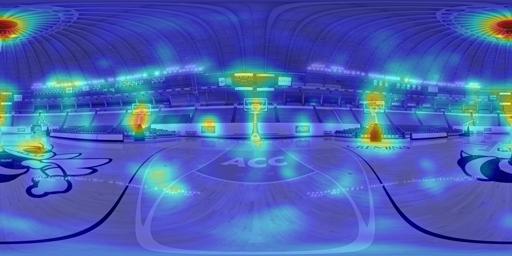}
	
	%\vspace{-1mm}
	\caption{\textbf{An example of a center-bias-free saliency map for a panoramic image collected by our system.} We uniformly sampled overlapping views from the panorama, projecting each to a regular photo perspective, and collected gaze data for each projection. We then projected the saliency maps back onto the panorama and averaged saliency for overlapping areas. This gives a panoramic saliency map with no photographer bias.}
	\label{fig:PanoSal}
\end{figure}

\textbf{Cost analysis and subject statistics.} We published the experiment on AMTurk and paid \$0.4 for a 4-min ‘Angry Bird’ task and \$0.05 for a 30-sec ‘Whac-a-mole’ task(resulting in an hourly payment of \$6). For an experiment with 20 images in the free viewing session, we are able to gather 300 images for an one-hour of work and pay \$0.02/image. Figure \ref{fig:TurkerStatistics} shows the basic demographic statistics and the distribution of geolocation of 200 randomly-sampled subjects who participated in our task.

%%%%%%%%%%%
% Evaluation %
%%%%%%%%%%%

\section{Evaluation}
We evaluated the performance of our system by: 1) the accuracy of gaze prediction, and 2) the quality of fixation estimation and saliency maps.

\begin{figure}[t]
	\begin{center}
		%\fbox{\rule{0pt}{2in} \rule{0.9\linewidth}{0pt}}
		\includegraphics[width=\linewidth]{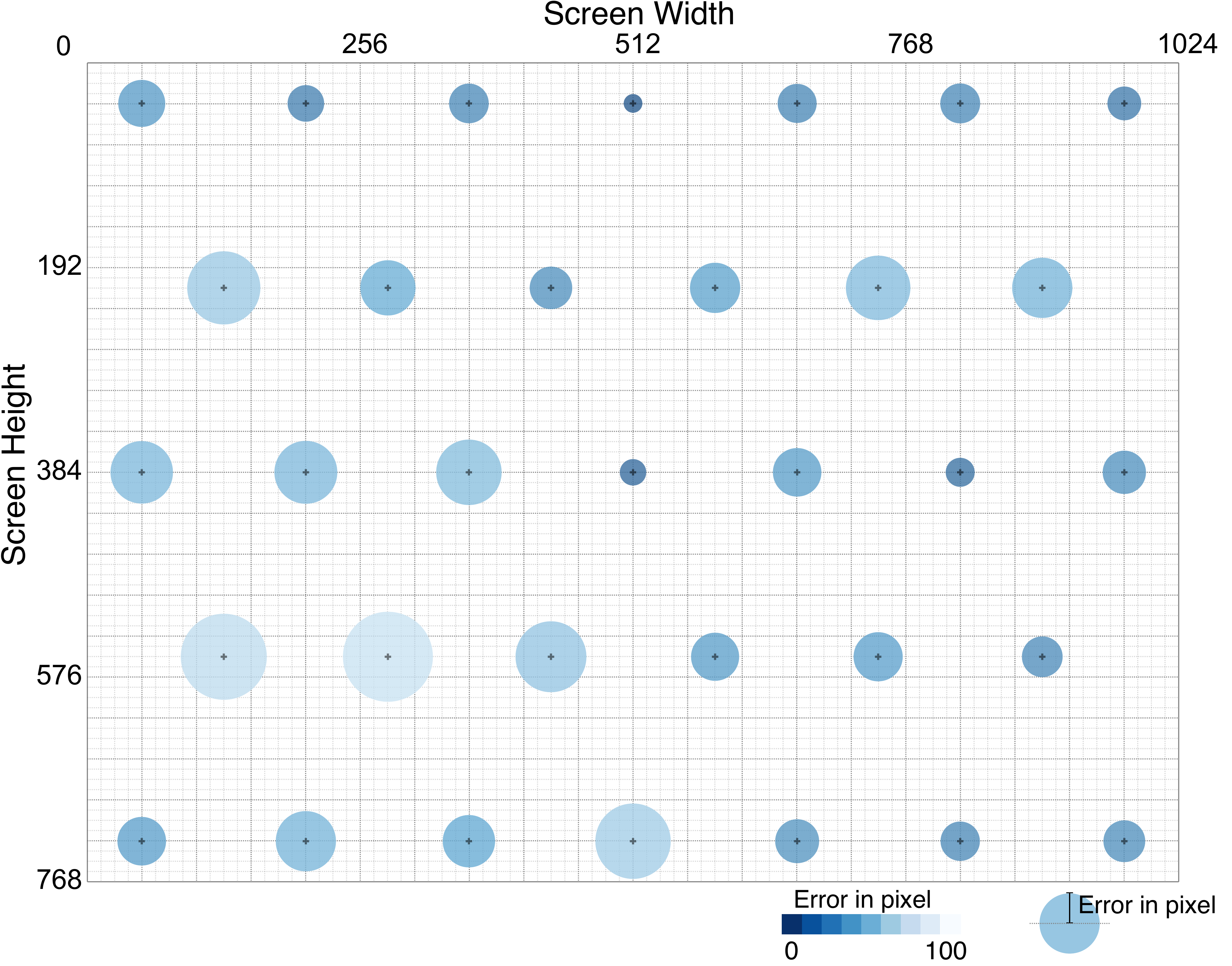}
	\end{center}
	\vspace{-4mm}	
	\caption{{\bf Gaze prediction accuracy evaluated by a 33-point chart.} The radius of each dot indicates the median error of the corresponding testing position of one experiment for one subject.}
	\vspace{-2mm}  
	\label{fig:long}
	\label{fig:onecol}
	\label{fig:point_positionerr}
\end{figure}

\begin{figure}[t]
	\begin{center}
		\includegraphics[width=\linewidth]{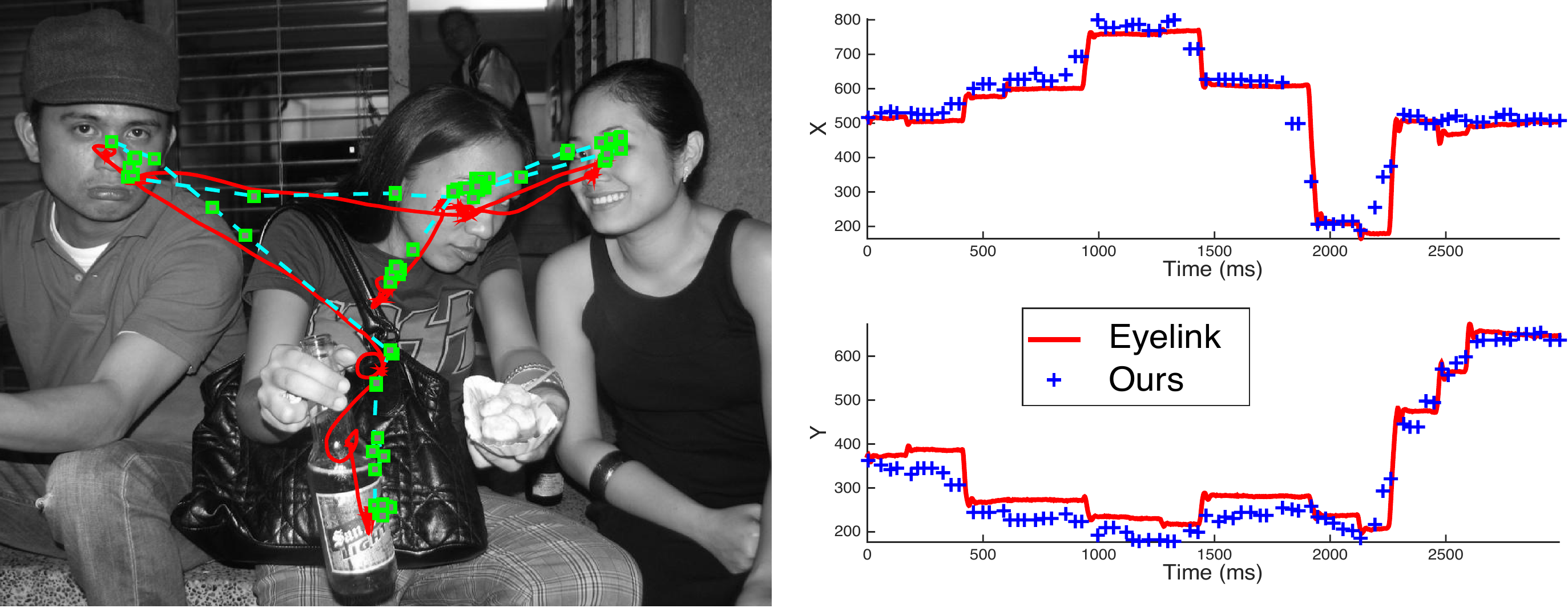}
	\end{center}
	
	\vspace{-2mm}
	\caption{{\bf Gaze prediction error during image free viewing.} On the left, the red
curve shows the eye position recorded by the Eyelink eye tracker operating at
1000 Hz, and the green squares indicate the predictions of our system with a 30
Hz sample rate. On the right is predicted gaze in the $x$ and $y$ direction over
time.}
	\vspace{-5mm}
	\label{fig:freeview_positionerr}
\end{figure}

\subsection{Gaze prediction accuracy}
We evaluated the accuracy of gaze prediction by asking users to perform two tasks: 1) gaze at uniformly sampled positions on screen; 2) free view natural images. Three observers (including an author) performed the tasks while being simultaneously tracked by our algorithm (using a Logitech HD C270 webcam) and a commercial eyetracker (Eyelink 1000). Users were seated in a headrest 56 cm away from a 26.5 $\times$ 37 cm CRT monitor with a resolution of 768 $\times$ 1024 pixels and refresh rate of 100 Hz. Nine-point calibration and validation were performed at the start of the session and after every two blocks of hits; mean error was $0.33\,^{\circ}$ (9.1 pixels).

To systematically evaluate the accuracy of different locations on the screen, we uniformly sampled 33 positions. Each point was shown for 1.5 seconds, and the last 0.5 second was used for evaluation. Figure \ref{fig:point_positionerr} illustrates the result tested on one subject, the radius indicates the median error by our approach. The median error is $1.06\,^{\circ}$ which is comparable to the state-of-the-art webcam based gaze prediction algorithm \cite{Lu11}.

Because different stimuli and tasks trigger different viewing behaviors, in addition to testing on fixed positions on the screen, we also compared the scan path predicted by our system with the one provided by the Eyelink 1000. Figure~\ref{fig:freeview_positionerr} illustrates the estimated errors on both $x$ and $y$ directions when a subject was freely viewing an image from Judd dataset. The sample rate of the web camera is set to 30fps, while the Eyelink 1000 samples gaze position at 1000 Hz, so we interpolated the tracking results from the Eyelink and calculated the error at time stamps when the webcam based gaze prediction is available. The average error is $1.32\,^{\circ}$ across 4 images from Judd dataset and three subjects.

\begin{figure}[t]
\begin{center}
%\fbox{\rule{0pt}{2in} \rule{0.9\linewidth}{0pt}}
\includegraphics[width=1\linewidth]{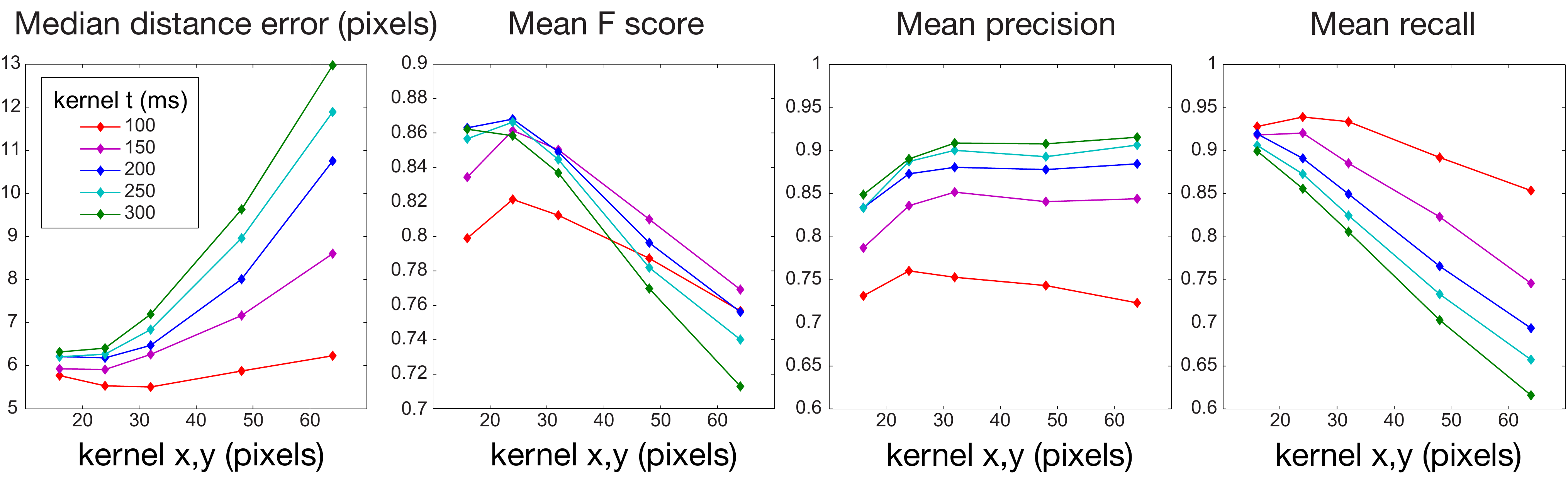}
\end{center}

\vspace{-3mm}
   \caption{\textbf{Performance of the meanshift fixation detection algorithm}. We ran the proposed fixation detection algorithm on 1000 randomly-sampled Judd images, for various kernel sizes, and evaluated it using the median position error, mean precision, recall and F-score.}
\vspace{-2mm}  
\label{fig:meanshift_kernel}
\end{figure}

\begin{figure}[t]
\begin{center}
%\fbox{\rule{0pt}{2in} \rule{0.9\linewidth}{0pt}}
   \includegraphics[width=1\linewidth]{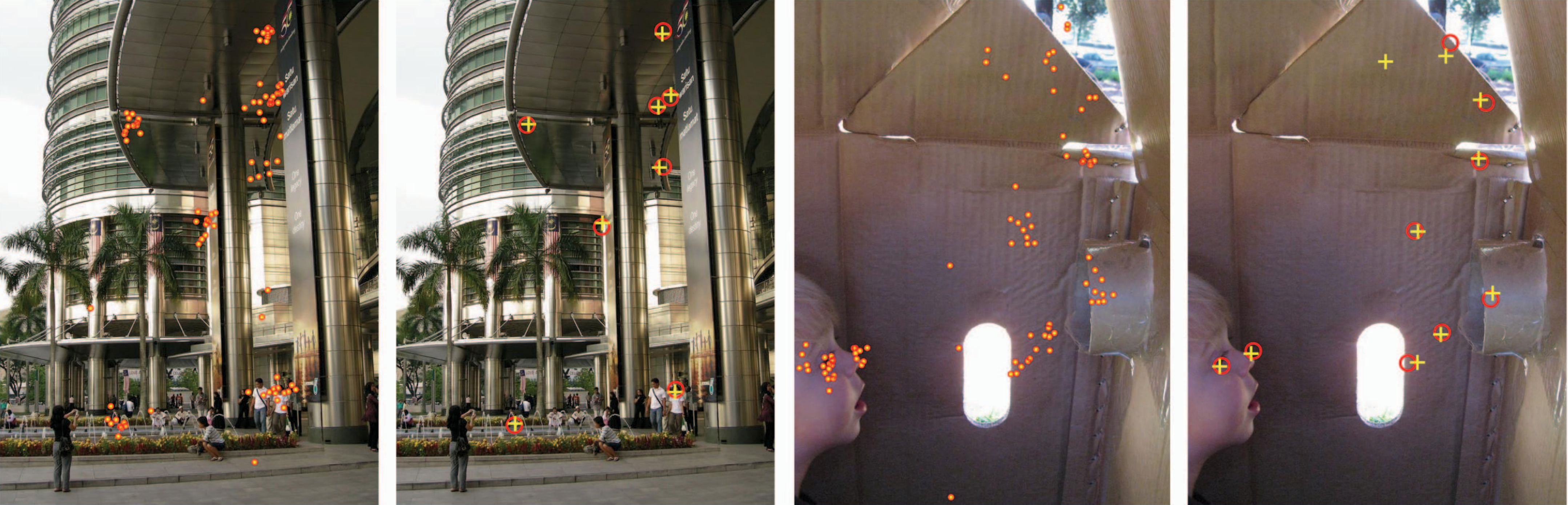}
\end{center}
\vspace{-3mm}
   \caption{{\bf Meanshift fixation detection results on two Judd images.} On left is the noisy, subsampled gaze data used for testing. On right are the detected fixations. Red circles indicate ground truth (detected in 240Hz data using code provided by Judd \etal.) and yellow crosses indicate fixations detected in the noisy 30Hz data using our meanshift approach.}
\vspace{-2mm}   
\label{fig:meanshift_fixation}
\end{figure}

\subsection{Fixation estimation and saliency maps}
\begin{figure}[t]
\begin{center}
%\fbox{\rule{0pt}{2in} \rule{0.9\linewidth}{0pt}}
   \includegraphics[width=0.8\linewidth]{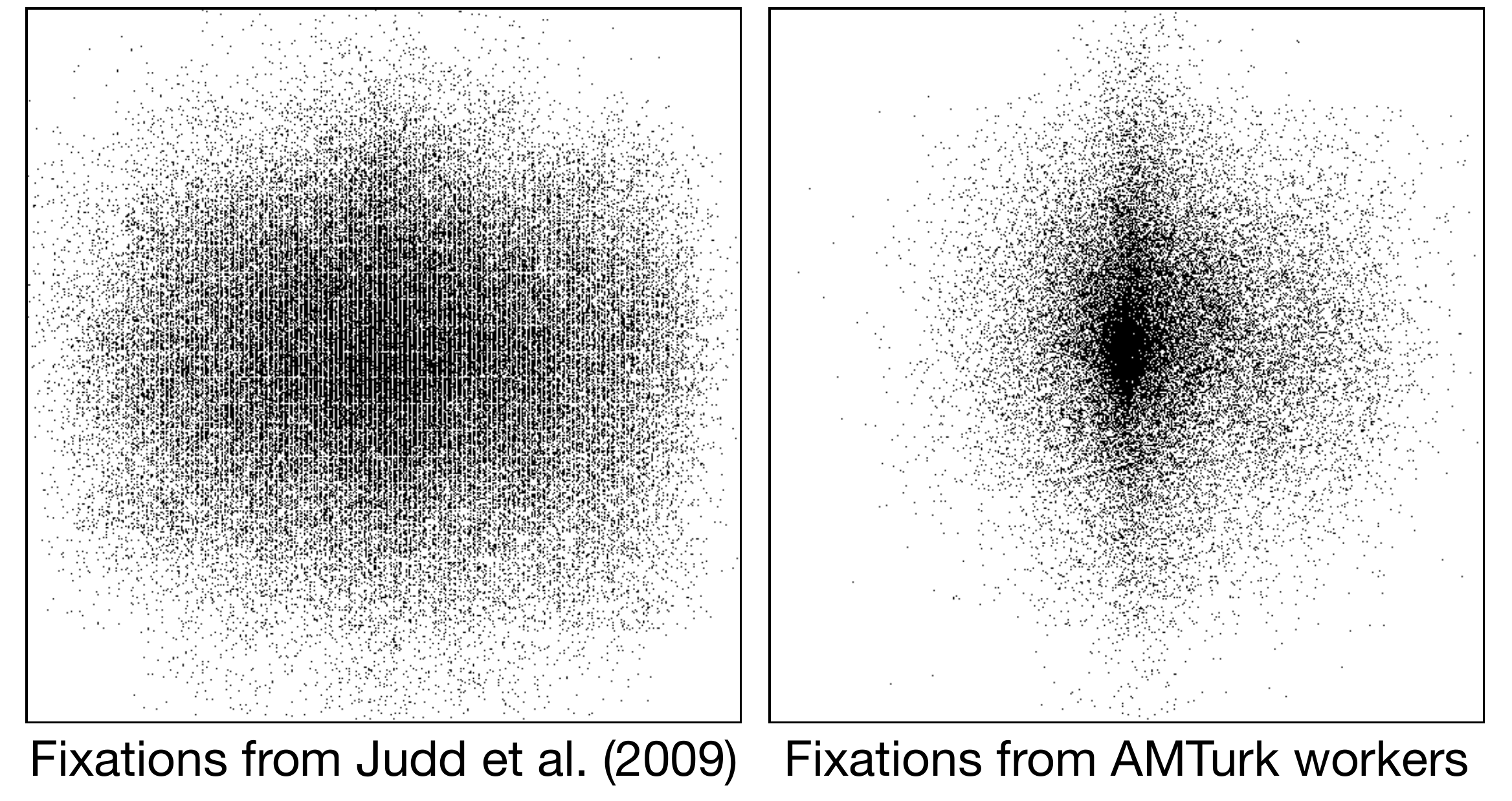}
\end{center}
\vspace{-4mm}
   \caption{\textbf{Distribution of fixations collected in the lab and on AMTurk.} Each distribution consists of 10,000 randomly-selected fixations on Judd images (the initial, central fixations were not included). Fixations captured by our method tended to be more central than those captured in the traditional setup.}
\vspace{-2mm}   
\label{fig:all_fixation}
\end{figure}

Raw gaze data includes fixations (stable gaze on one point in the image) and saccades (rapid eye movements from one fixation point to the next). Typically, only fixation points are used to test saliency models. We evaluated our data quality by using the raw gaze data from Judd \etal \cite{Judd09}. We selected 1000 random subject/image pairs, and obtained ground truth fixation locations using their published code. We then simulated webcam tracking by subsampling their data at 30 Hz and adding Gaussian position noise ($\sigma = 10$ pixels). We extracted fixations from the 30 Hz data using our meanshift algorithm and compared our detection results to the ground truth fixations. The results on two Judd Images are presented in Figure~\ref{fig:meanshift_fixation}. Median position error, mean precision, recall, and F-score for various kernel sizes are also shown in Figure \ref{fig:meanshift_kernel}. Our proposed algorithm provided a satisfactory estimation of fixations. Furthermore, we compared the distribution of fixations presented in ~\cite{Judd09} with the one we collected on AMTurk (Figure~\ref{fig:all_fixation}).

\begin{figure}[t]
\begin{center}
%\fbox{\rule{0pt}{2in} \rule{0.9\linewidth}{0pt}}
   \includegraphics[width=1\linewidth]{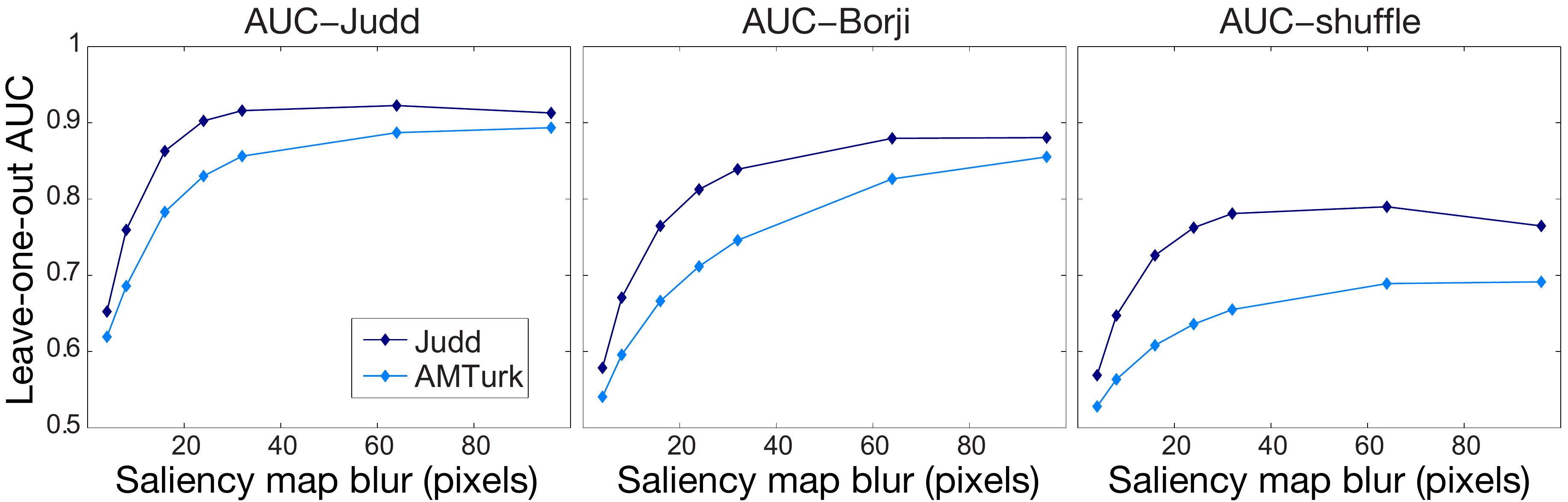}
\end{center}
\vspace{-2mm}
\caption{{\bf Leave-one-out AUC for saliency prediction}. We predicted a single Judd or AMTurk subject's fixations with a ground truth saliency map built from 14 Judd subjects' fixations.}
\vspace{-4mm}
\label{fig:leave_one_out}
\end{figure}

We also evaluated the agreement between the fixation data from our system and the ground truth fixation data on the Judd dataset. For each image, we computed a ground truth saliency map using fixation data from 14 of the 15 subjects, then calculated the leave-one-out AUC by predicting the remaining Judd subject or a single AMTurk subject. The result is shown in Figure~\ref{fig:leave_one_out}. We repeated this process for all of the AMTurk subjects on each image (average 6.7 subjects/image), using a different random Judd subject for each comparison. We removed the initial, central fixation on each image. Average AUC for various levels of map blur via various metrics are shown in Figure \ref{fig:leave_one_out}. In general, AMTurk subjects are not predicted by the ground truth saliency map as well as Judd subjects are. This may reflect the higher positional noise in the AMTurk data and demographic difference between the AMTurk and Judd subject pools.

\begin{figure}[t]
\begin{center}
%\fbox{\rule{0pt}{2in} \rule{0.9\linewidth}{0pt}}
   \includegraphics[width=0.8\linewidth]{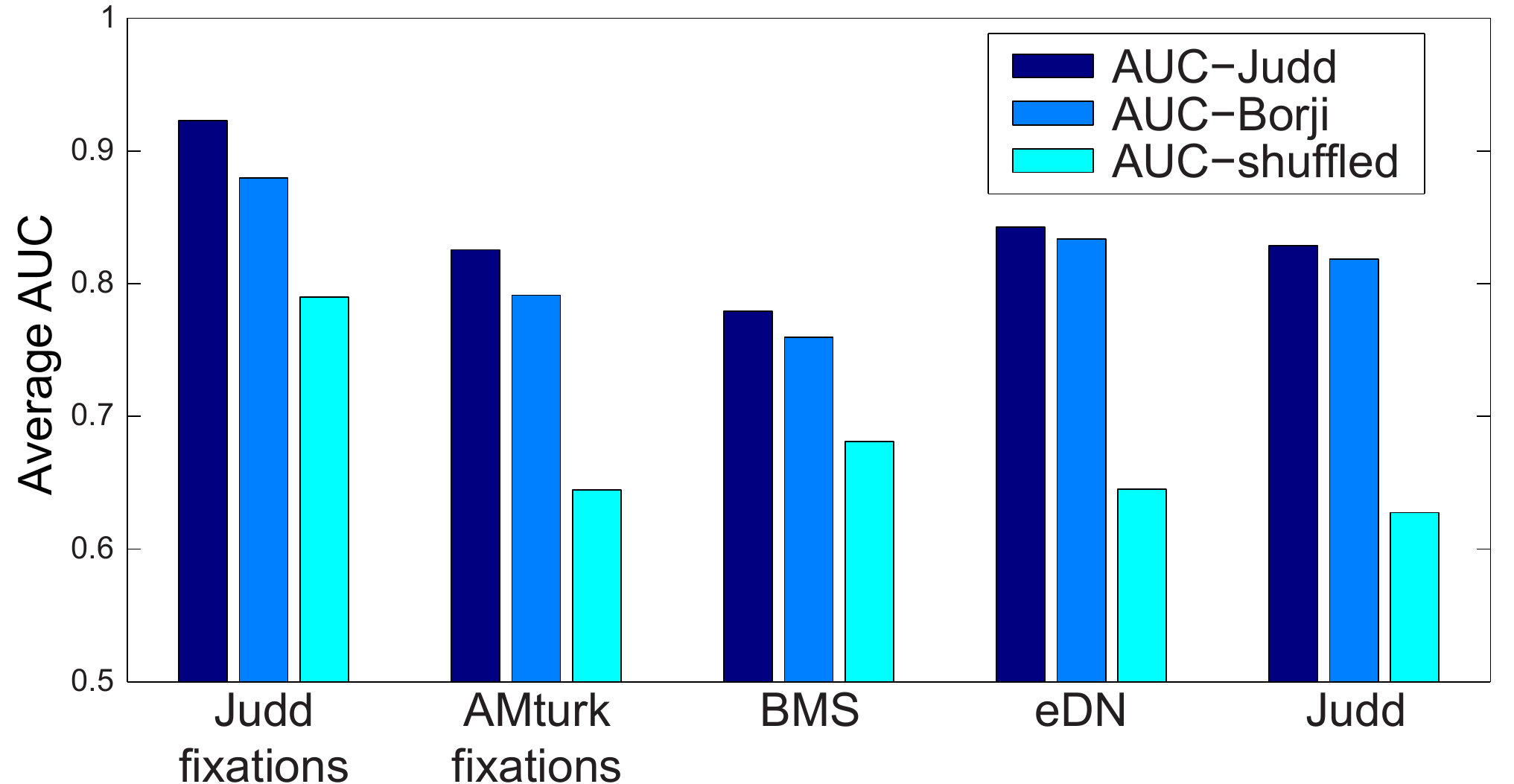}
\end{center}

\vspace{-3mm}
   \caption{{\bf Average AUC for saliency prediction.} We compared the saliency map based on AMTurk subjects' fixations to various state-of-the-art computer models of saliency\cite{Vig14,Zhang13,Judd09} and the Judd inter-subject-agreement baseline.}
\vspace{-1mm}
\label{fig:AUC_comparison}
\end{figure}

We further used the AMTurk fixation data as a saliency map to predict the 15 Judd subjects and compared these results with the state-of-the-art saliency algorithms. As shown in Figure \ref{fig:AUC_comparison}, the AMTurk saliency map is similar to top performing models by various AUC metrics.

%%%%%%%%%
% Dataset %
%%%%%%%%%
\section{iSUN: A large dataset for natural scenes}

\begin{figure}[t]
\includegraphics[width=1\linewidth]{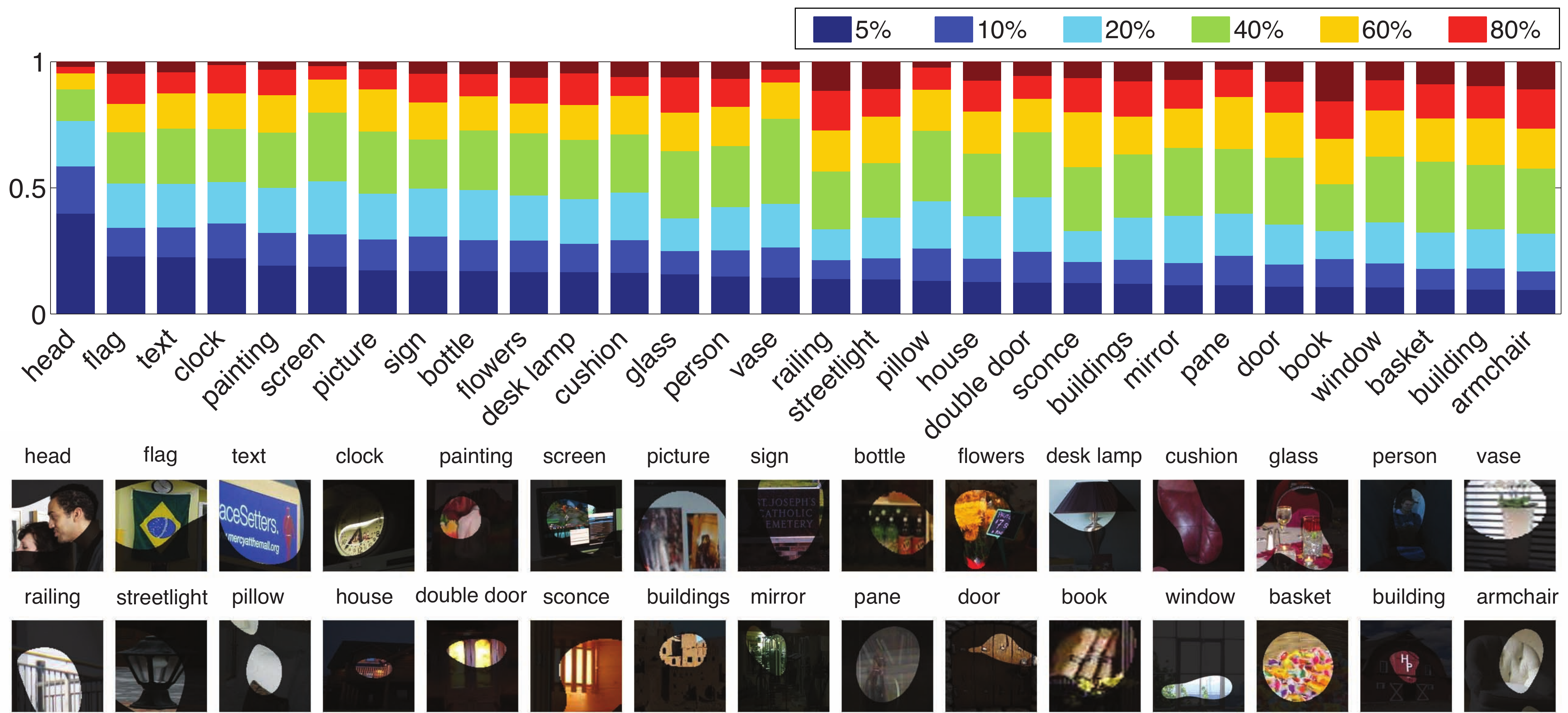}
\caption{\textbf{Object saliency statistics}. For each image, we threshold the saliency map to create a binary map that selects N\% of the image. We then compute the average overlap between the thresholded saliency map and a binary mask for a particular object category. For example, if there is a head in an image, on average the head area will have around 0.45 overlap with the top 5\% most salient image area. The bottom rows show example objects cropped from our images, with a binary mask representing the top 5\% most salient region.}
\vspace{-1mm}
\label{fig:salientObject}
\end{figure}

We used our tool to construct a large-scale eye tracking dataset. We selected
natural scene images from the SUN database~\cite{Xiao10} --  the standard dataset for
scene understanding. We selected 20,608 images from SUN database with full
object/region annotations and recorded eye tracking data from subjects free
viewing these images. At the time of submission, we have obtained saliency maps
for all images, with an average of 3 observers per image. Table \ref{table:compDataset} shows a brief
comparison between iSUN and several other eye tracking datasets, and Figure \ref{fig:salientObject} shows the top 30 most salient object categories. We will continue collecting data on more images with the same protocol.

\begin{table}[t]
\setlength{\tabcolsep}{1pt}

\scriptsize
\begin{center}
\begin{tabular}{c | c c c c c c}
%\begin{tabular}{c|c|c|c|c|c|c}
\hline
Dataset & Images & Subjects & Semantics & Categories & Objects  & Annotaions\\
\hline
iSUN & 20608 & $3$& scene & 397 & 2333 classes & Fully  \\
\hline
SalObj\cite{Li14} & 850 & 12& object & - & 1296 & Fully\\
\hline
MIT\cite{Judd09} &  1003 & 15 & Flickr \& LabelMe & 2 & -&-\\
\hline
NUSEF\cite{Ramanathan12} & 758 & 25.3  & human \& reptile %Expressive face, nude, action, reptile and affect-variant group 
&  6 & - &-\\
\hline
Toronto\cite{Bruce09} & 120 & 20& scene& 2 & - &-\\
\hline
Koostra\cite{Kootstra08} &  101& 31& - & 5 & -&-\\
\hline
%\end{tabularx}
\end{tabular}
\end{center}

\vspace{-3mm}
\caption{\textbf{Comparison between iSUN and several other free-viewed image datasets.} Our iSUN database is by far the largest in terms of number of images. The iSUN images are fully-annotated scenes from the SUN database.}
\vspace{-1mm}
\label{table:compDataset}
\end{table}

\section{Conclusions}
In this paper, we present a webcam based system for crowdsourced eye tracking data collection from Amazon Mechanical Turk. We demonstrate its effectiveness by comparing it to the gaze data obtained from commercial eye trackers. We use our approach to obtain free-viewed eye tracking data for a large number of natural scene images. Our system can be  also easily generalized to various types of visual stimuli and tasks for eye tracking experiments. We believe that the proposed system will be a useful tool for the research community to collect large-scale eye tracking data on a variety of tasks.

\newpage
\paragraph{Acknowledgement.}
This work is supported by gift funds from Google, MERL, and Project X to JX.
\vspace{-7mm}
{\small
\bibliographystyle{ieee}
\bibliography{iccvref}
}
\end{document}